\documentclass{l4dc2025}
\usepackage{hyperref}
\usepackage{amsmath}  % improve math presentation
\usepackage{amssymb}
\usepackage{mathtools}
\usepackage{caption} % figure and subfigure caption
\usepackage{float}
\usepackage{comment}
\usepackage{subcaption}
\usepackage{algorithm,algcompatible,algpseudocode}
\usepackage{indentfirst} % Indentation in the first paragraph
\newtheorem*{assumption*}{\assumptionnumber}
\providecommand{\assumptionnumber}{}
\makeatletter

\usepackage{amsfonts} % improve math presentation
\usepackage{graphicx} % takes care of graphic including machinery

\DeclareMathOperator*{\argmin}{argmin}

\DeclareMathOperator*{\E}{\mathbb{E}}

\newcommand{\abs}[1]{\left|#1\right|}

\newcommand{\brk}[1]{\left(#1\right)}
\newcommand{\Brk}[1]{\Big(#1\Big)}
\newcommand{\bsq}[1]{\left[#1\right]}
\newcommand{\bcur}[1]{\left\{#1\right\}}

\newcommand{\prob}[1]{\mathbb P\brk{#1}}

\newcommand{\ceil}[1]{\lceil#1\rceil}

\newcommand{\C}{\mathcal{C}}
\newcommand{\D}{\mathcal{D}}
\renewcommand{\P}{\mathcal{P}}
\newcommand{\I}{\mathcal{I}}
\newcommand{\X}{\mathcal{X}}
\newcommand{\Y}{\mathcal{Y}}

\newcommand{\RR}{\mathbb{R}}
\newcommand{\PP}{\mathbb{P}}
\DeclareMathAlphabet{\mymathbb}{U}{BOONDOX-ds}{m}{n}

\newcommand{\fullpaper}{the technical report (\url{https://bit.ly/3CQ3jpO})}

\newcommand{\videoInD}{\url{https://bit.ly/3CGKB3C}}

\newcommand{\videoOOD}{\url{https://bit.ly/4eNK5ym}}

\def\arxiv{1} % for arxiv version
\newtheorem{rmk}{Remark}
\newtheorem{prop}{Proposition}
\if\arxiv0
\newcommand{\propRmkSpace}{\vspace{-0.5em}}
\newcommand{\itemSpace}{\vspace{-0.3em}}
\fi
\if\arxiv1
\newcommand{\propRmkSpace}{}
\newcommand{\itemSpace}{}
\fi
\title[Safe ACC Under Perception Uncertainty]{Safe Adaptive Cruise Control Under Perception Uncertainty: A Deep Ensemble and Conformal Tube Model Predictive Control Approach}
\usepackage{times}
\coltauthor{
\Name{Xiao Li} \Email{hsiaoli@umich.edu}\\
\Name{Anouck Girard} \Email{anouck@umich.edu}\\
\Name{Ilya Kolmanovsky} \Email{ilya@umich.edu}\\
\addr University of Michigan, Ann Arbor, MI, USA
}

\if\arxiv0

\usepackage{titlesec}
\titlespacing*{\section}{0pt}{0.7\baselineskip}{0.2\baselineskip}
\titlespacing*{\subsection}{0pt}{0.5\baselineskip}{0.2\baselineskip}
\fi
\begin{document}
%=================================================================
\if\arxiv0
% Change space above and below equations
\setlength{\abovedisplayskip}{3pt}
\setlength{\belowdisplayskip}{3pt}
\setlength{\abovedisplayshortskip}{3pt}
\setlength{\belowdisplayshortskip}{3pt}
%  and space after figure captions
\setlength{\belowcaptionskip}{-5pt}
\setlength{\abovecaptionskip}{-8pt}
\fi
%=================================================================
\maketitle

\if\arxiv0
\vspace{-2em}
\fi
\begin{abstract}%
Autonomous driving heavily relies on perception systems to interpret the environment for decision-making. To enhance robustness in these safety critical applications, this paper considers a Deep Ensemble of Deep Neural Network regressors integrated with Conformal Prediction to predict and quantify uncertainties. In the Adaptive Cruise Control setting, the proposed method performs state and uncertainty estimation from RGB images, informing the downstream controller of the DNN perception uncertainties. An adaptive cruise controller using Conformal Tube Model Predictive Control is designed to ensure probabilistic safety. Evaluations with a high-fidelity simulator demonstrate the algorithm's effectiveness in speed tracking and safe distance maintaining, including in Out-Of-Distribution scenarios.
\end{abstract}

\begin{keywords}%
Perception Uncertainty, Deep Neural Network, Conformal Prediction, Model Predictive Control
\end{keywords}

% ==================================================
\section{Introduction}\label{sec:intro}
Advances in Deep Neural Networks (DNNs) have significantly enhanced visual-based perception capabilities in autonomous driving. However, as black-box models, DNNs lack interpretability and operate in a high-dimensional image space vulnerable to pixel-level attacks~\citep{goodfellow2016deep}, and to Out-Of-Distribution (OOD) observations, which are unseen in training data~\citep{yang2024generalized}. The integration of these models can impact the autonomous driving systems' safety through other system modules, such as control and decision-making~\citep{zablocki2022explainability}.

Researchers have been investigating Bayesian frameworks to model inference uncertainty in DNNs~\citep{neal2012bayesian, mackay1992practical}, despite their computational intensity. Bayesian DNNs are attractive as they offer theoretical guarantees useful in uncertainty quantification by assuming prior distributions on model parameters or architectures.
To address computational demands, the Monte Carlo Dropout method~\citep{gal2016dropout} has been proposed that approximates prior distributions through empirical dropout sampling. Similarly, Deep Ensembles~\citep{deep_ensemble} leverage a diverse set of DNN architectures, enabling a voting mechanism that enhances inference robustness against OOD and adversarial attacks at the pixel level.
Other approaches, such as Laplace Approximation~\citep{ritter2018scalable} and Variational Neural Networks~\citep{graves2011practical}, have also been developed. A comprehensive review is available in \cite{abdar2021review}. While these methods are easy to implement, they may lack theoretical guarantees.
In contrast, Conformal Prediction~\citep{conformal_l_vovk} offers distribution-free uncertainty quantification, relying on exchangeability assumptions. This work combines Deep Ensembles with Conformal Prediction for robust uncertainty quantification with statistical guarantees, and it demonstrates how such guarantees can be integrated into Model Predictive Control (MPC) using Adaptive Cruise Control (ACC) as an application.

Additionally, methods have been explored to control robots using DNN-based perception and prediction~\citep{tong2023enforcing} integrated with Control Barrier Functions, though they do not address the uncertainties inherent in DNN perception. 
In autonomous driving, researchers have developed DNN architectures to quantify classification uncertainties~\citep{liu2022pnnuad} and trajectory prediction uncertainties~\citep{ding2021capture, tang2022prediction}. However, these methods only provide heuristic uncertainty estimates without theoretical support.
Control co-design approaches have also been investigated, assuming bounded DNN errors to achieve trajectory tracking~\citep{dean2020robust} and ensure system-level safety~\citep{li2024system}. These approaches, however, are limited to in-distribution scenarios~\citep{dean2020robust}. 
Previous studies~\citep{li2024autonomous} have applied Gaussian assumptions to model perception uncertainties in DNNs, though this assumption may lack practical validity in complex environments.
Conformal Prediction has been previously integrated into MPC design in \cite{lindemann2023safe}. Differently, in this paper, we consider perception uncertainties; their handling requires us to propagate the uncertainty of Conformal Prediction results through the system dynamics.  

More specifically, in this paper, we integrate Deep Ensembles with the Conformal Prediction to quantify uncertainty associated with image-based sensory information. We illustrate the use of this uncertainty quantification in a Conformal Prediction Tube MPC to control vehicle acceleration in an ACC application. The contributions of this paper are summarized as follows:
\itemSpace
\begin{itemize}\setlength\itemsep{0em}
    \item We develop a Deep Ensemble that assembles a heterogeneous set of DNN architectures to enhance robustness in state estimation from images and provide reliable uncertainty measures. The DNNs are optimized through iterative pruning and fine-tuning to reduce memory demand.
    
    \item We integrate the Deep Ensemble with the Conformal Prediction, transforming heuristic uncertainty estimates into statistically guaranteed uncertainty measures under exchangeabilty assumptions. This integration demonstrates reliable uncertainty quantification, even in the presence of pixel-level adversarial attacks and OOD observations.
    
    \item We design a Conformal Tube MPC that leverages the uncertainty estimates for uncertainty-aware trajectory prediction as a tube of Conformal Prediction sets. The proposed MPC is formulated as a Quadratic Program, ensuring recursive feasibility with a probabilistic lower-bound guarantee of system safety.
\end{itemize}
\itemSpace

This paper is organized as follows: In Sec.~\ref{sec:problem}, we introduce the ACC problem, along with model assumptions and design objectives. In Sec.~\ref{sec:method}, we present our development of a Deep Ensemble and Conformal Prediction approach for state estimation, which subsequently informs a Conformal Tube MPC to control the acceleration of the ego vehicle. In Sec.~\ref{sec:result}, we demonstrate that our method provides robust state estimation and quantifies uncertainty in the presence of adversarial attacks and OOD observations. Additionally, we validate the proposed adaptive cruise controller using a high-fidelity simulator. Finally, Sec.~\ref{sec:conclusion} presents a summary and a direction for future work.
% ==================================================
\section{Problem Formulation}\label{sec:problem} 
\begin{figure}[!h]
\begin{center}
\includegraphics[width=0.8\linewidth]{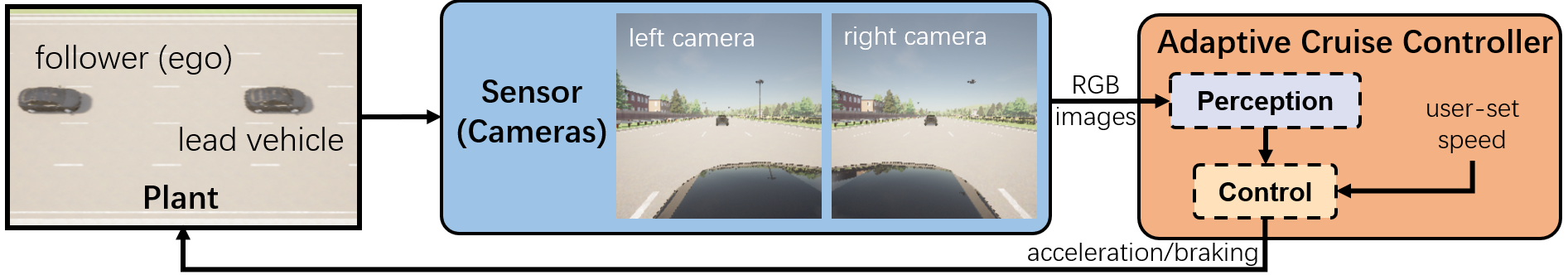}
\end{center}
\caption{A schematic diagram of the Adaptive Cruise Control implementation.}
\label{fig:problem}
\end{figure}

In this study, we investigate an ACC scenario where the ego vehicle follows a lead vehicle (see Fig.~\ref{fig:problem}). Our focus is on integrating visual-based perception and control mechanisms. The ego vehicle uses cameras to capture RGB images, which are processed to estimate system states. The Adaptive Cruise Controller then commands the ego vehicle to accelerate or brake to maintain a safe distance while adhering to a user-specified speed. This scenario presents challenges in both perception and control aspects.

For perception, we assume two cameras are mounted on the left and right sides at the front of the ego vehicle to emulate stereo vision. The perception is modeled using the following mapping,
\begin{equation}\label{eq:model:percept}
    x_k = y_k\brk{\bcur{I_{l,k-i},I_{r,k-i}}_{i=0}^{M-1}} + e_k\brk{\bcur{I_{l,k-i},I_{r,k-i}}_{i=0}^{M-1}}, 
\end{equation}
where $x_k=[d_k~ \Delta v_k ~ v_k]^T\in\RR^3$ is the state vector;
$d_k\in \RR^{\geq 0}$ is the distance headway defined as the bumper-to-bumper distance in meters between the ego and the lead vehicle;
$\Delta v_k = v_k^{(l)} - v_k \in\RR$ in $\textrm{m/s}$ is the difference between the lead vehicle speed $v_k^{(l)}$ and ego vehicle speed $v_k$;
$I_{l,k},I_{r,k}\in\I \subset\RR^{3\times 224\times 224}$ are RGB images of $3$ color channels and of size $224\times 224$ acquired from the left and right cameras, respectively;
$y_k\in\RR^3$ denotes the state estimation derived from a historical image buffer $\bcur{(I_{l,k-i},I_{r,k-i})}_{i=0}^{M-1}$ of length $M$; $e_k\in\RR^3$ is the unknown state estimation error and follows an unknown distribution.
Image-based perception has been focused on enhancing estimation accuracy by minimizing $e_k$ using Convolutional Neural Networks (CNNs).
However, $e_k$ is influenced by inputs residing in a high-dimensional image space $\I$, making it highly unpredictable when subjected to disturbances at the pixel level, such as adversarial attacks in \cite{goodfellow2016deep}. For enhanced safety, our perception developments aim to achieve the following objectives:
\begin{itemize}\setlength\itemsep{0em}
    \item \textit{Estimation Accuracy:} improve estimation accuracy by minimizing the estimation error $e_k$.
    \item \textit{Uncertainty Quantification:} provide a reliable quantification of this unknown error.
\end{itemize}

For ego vehicle control, we use the following discrete-time model to represent the kinematics,
\begin{equation}\label{eq:model:kine}
    x_k = A x_k + B a_k,
    ~ 
    A = 
    \scalebox{0.8}{$
    \begin{bmatrix}
        1 & \Delta t & 0\\
        0 & 1 & 0\\
        0 & 0 & 1
    \end{bmatrix},
    $}
    ~ B = 
    \scalebox{0.8}{$
    \begin{bmatrix}
        -\frac{1}{2} \Delta t^2 \\
        - \Delta t \\
        \Delta t
    \end{bmatrix}
    $}
\end{equation}
where $a_k \in \RR$ is the acceleration/deceleration of the ego vehicle in $\textrm{m/s}^2$, and $\Delta t > 0$ is the time elapsed in seconds between discrete time instances $t_k$ and $t_{k+1}$.
Since we focus on car-following, we consider only the longitudinal vehicle kinematics in \eqref{eq:model:kine}.
In deriving \eqref{eq:model:kine}, we assume that the lead vehicle maintains a constant speed between the time instances $t_{k-1}$ and $t_k$. 
We focus on a Receding-Horizon-based optimization scheme that optimizes a sequence of controls $\brk{a_{k+i}}_{i=0}^{N-1}$ over a horizon of length $N$ while adhering to the following objectives and constraints:
\begin{itemize}\setlength\itemsep{0em}
    \item \textit{Safety}: keep an adequate distance headway $d_k$ to the lead vehicle to prevent collisions.
    \item \textit{Fuel Economy}: minimize the accumulated acceleration effort $\sum_{i=0}^{N-1} \abs{a_{k+i}}$.
    \item \textit{Driving Comfort}: minimize the rate of change in the acceleration trajectory $(a_{k+i})_{i=0}^{N-1}$.
    \item \textit{Speed Tracking}: track the driver-set speed $v_s$.
    \item \textit{Speed and Control Limits}: keep the speed $v_k\in[v_{\min}, v_{\max}]$ and acceleration $a_k\in[a_{\min}, a_{\max}]$.
    \item \textit{Uncertainty Awareness:} integrate perception results with control development so that the control system is aware of perception uncertainties, enhancing overall ACC system safety.
\end{itemize}
% ==================================================
\section{Method}\label{sec:method}
\begin{figure}[t!]
\begin{center}
\includegraphics[width=0.99\linewidth]{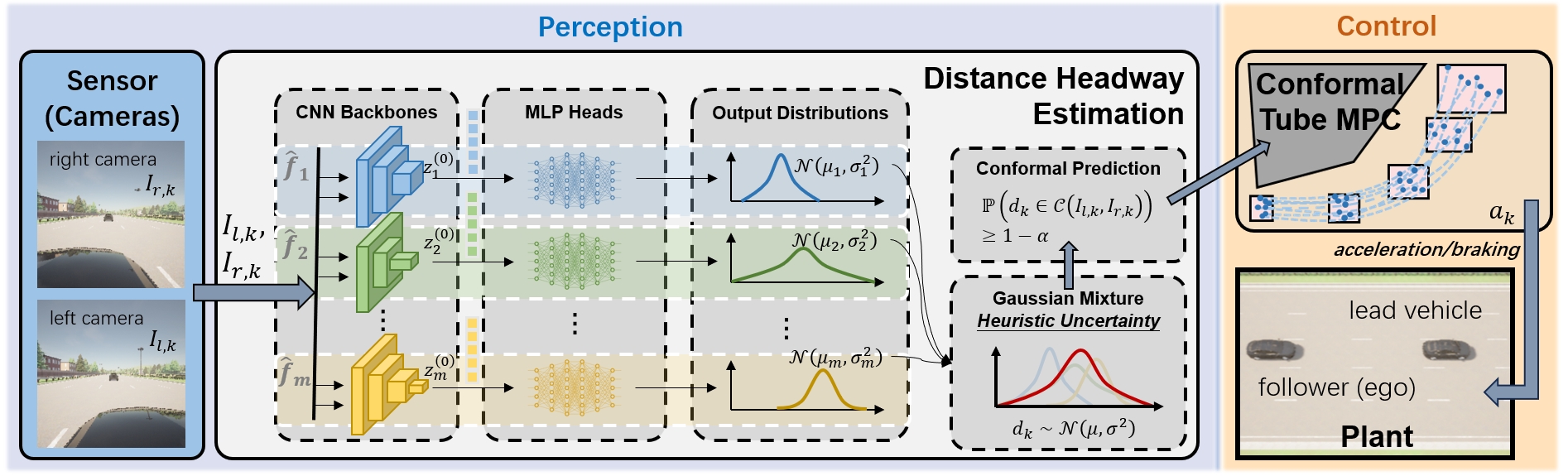}
\end{center}
\caption{Adaptive cruise controller pipeline: The Deep Ensemble uses a diverse set of DNNs for estimation, incorporating uncertainty quantification via Conformal Prediction. The Perception Uncertainty-Aware Conformal Tube MPC then computes the acceleration commands.}
\label{fig:method}
\end{figure}

As shown in Fig.~\ref{fig:method}, we propose an adaptive cruise controller that employs a Deep Ensemble to estimate system states with formal uncertainty quantification via the Conformal Prediction. The concept of Conformal Prediction is introduced in Sec.~\ref{subsec:method:conformal}, and the perception solution is detailed in Sec.~\ref{subsec:method:perception}. The estimation and uncertainty quantification is then integrated into a Conformal Tube MPC design for vehicle acceleration control, as discussed in Sec.~\ref{subsec:method:control}.
\if\arxiv0 
The proofs of theoretical results are presented in \fullpaper.
\fi

% ==================================================
\subsection{Preliminaries: Conformal Prediction}\label{subsec:method:conformal} 
We consider regression problems aiming to learn an unknown mapping $f: \X \to \Y$ from input domain $\X$ to output domain $\Y$. We denote the training, testing, and calibration datasets as $\D_\textrm{train}$, $\D_\textrm{test}$, and $\D_\textrm{cali}$, respectively, where each data point is a pair $(X_i, Y_i)$ with $X_i \in \X$ and $Y_i = f(X_i)$. We present Algorithm~\ref{al:conformal} and Theorem~\ref{thm:conformal} to establish probabilistic coverage for prediction results. A tutorial is available in \cite{conformal_tutorial}. 

\begin{algorithm}[!h]
\caption{(Split) Conformal Prediction~\citep{conformal_l_vovk}}
\label{al:conformal}
\begin{algorithmic}[1]
\Require {A training set $\D_\textrm{train}$ and a calibration set $\D_\textrm{cali}=\bcur{X_i,Y_i}_{i=1}^n$; prediction model $\hat{f}:\X\to\Y$; miscoverage rate $\alpha\in[0,1]$.}
\State train the prediction model $\hat{f}$ on the training set $\D_\textrm{train}$ and yield $\hat{f}_{\D_\textrm{train}}$.
\State define a score function $S(X,Y)=S(X,Y | \hat{f}_{\D_\textrm{train}})\in\RR$ that is negatively-oriented. 
\Comment{{\color{lightgray}\footnotesize Lower score implies better prediction performance of $\hat{f}_{\D_\textrm{train}}$. For example, if $\Y\subset \RR$, we can define $S(X,Y)=\abs{\hat{f}_{\D_\textrm{train}}(X)-Y}$.}}
\State calculate the scores for all points in $\D_\textrm{cali}$, i.e., $S_i=S(X_i,Y_i), i=1,\dots,n$.
\State identify conformal quantile $q(\alpha)=\inf\bcur{q:~ \abs{\bcur{i:~ S(X_i,Y_i)\leq q}}\geq \ceil{(n+1)(1-\alpha)}}$ 
\Comment{{\color{lightgray}\footnotesize
$\abs{A}$ is the cardinality of the input set $A$, $\ceil{\cdot}$ is the ceiling function and the quantile $q(\alpha)$ attains the value of the $\ceil{(n+1)(1-\alpha)}$ smallest score among $S_1,S_2,\dots,S_n$. In case of $\alpha <\frac{1}{1+n}$, we assume $q(\alpha)=\infty$. }}
\State define prediction set $\C(X)=\bcur{Y: ~ S(X,Y)\leq q(\alpha)}$
\\\Return{ prediction set $\C(X)$ } 
\end{algorithmic}
\end{algorithm}

\if\arxiv0
\vspace{-1em}
\fi
\begin{theorem}\label{thm:conformal}
    \citep{conformal_l_vovk} For arbitrary test point $(X_\textrm{test}, Y_\textrm{test})\in\D_\textrm{test}$, if $(X_i, Y_i)_{i=1}^n$ and $(X_\textrm{test}, Y_\textrm{test})$ are i.i.d., then the prediction set $\C(X_\textrm{test})$ produced by the Algorithm~\ref{al:conformal} provides a coverage of the true result $Y_\textrm{test}$ with a probability of at least $1-\alpha$, i.e., $\prob{Y_\textrm{test}\in \C(X_\textrm{test})}\geq 1-\alpha $.
\end{theorem}

\propRmkSpace
\begin{rmk}\label{rmk:conformal}
The independent and identically distributed (i.i.d.) assumption in Theorem~\ref{thm:conformal} can be relaxed to the exchangeability of $(X_i, Y_i)_{i=1}^n$ and $(X_\textrm{test}, Y_\textrm{test})$. We note that the i.i.d. assumption implies exchangeability of random variables, whereas exchangeability implies that random variables are identically distributed but not necessarily independent~\citep{kuchibhotla2020exchangeability}.   
\end{rmk}
\propRmkSpace

% ==================================================
\subsection{Deep Ensemble and Conformal Prediction}\label{subsec:method:perception} 

As presented in Sec.~\ref{sec:problem}, we focus on estimating the state vector $x_k$ while quantifying uncertainties from image inputs. This task is decomposed into three steps: First, we estimate and quantify uncertainty for the distance headway $d_k$ using a Deep Ensemble with images $I_{l,k}$ and $I_{r,k}$ from the left and right cameras. Secondly, we apply the Conformal Prediction framework to provide statistical guarantees for the uncertainty estimates. Lastly, assuming the vehicle speed $v_k$ is known and given images $I_{l,k-1}$ and $I_{r,k-1}$ from previous time steps, we estimate the velocity state $\Delta v_k$.

At first, as illustrated in Fig.~\ref{fig:method}, we assemble $m$ different DNN paths denoted as $\hat{f}_i$, $i=1,\dots,m$. Each DNN path in parallel processes the input RGB images $I_{l,k}$ and $I_{r,k}$ into two scalar outputs $\mu_i\in\RR,\sigma_i^2\in\RR^+$, $i=1,\dots,m$ according to the following equations,
\begin{subequations}\label{eq:subnet}
\begin{equation}
     [\mu_i~\sigma_i^2]^T = \hat{f}_i(I_{l,k}, I_{r,k}) = \hat{f}_i^\textrm{MLP} \circ \hat{f}_i^\textrm{CNN} (I_{l,k}, I_{r,k}),
\end{equation}
\begin{equation}\label{eq:subnet:mlp:final}
[\mu_i~\sigma_i^2]^T = \hat{f}_i^\textrm{MLP}(z^{(0)}) = W_i^{(\ell_i)}z^{(\ell_i)}+b_i^{(\ell_i)},
\end{equation}
\begin{equation}\label{eq:subnet:mlp:hidden}
    z^{(j+1)} = \sigma_{\text{ReLU}}\left(W_i^{(j)}z^{(j)}+b_i^{(j)}\right), 
    ~
    j = 0,\dots,\ell_i-1,
    ~~
    z^{(0)} = [z_{l,k}^T ~~ z_{r,k}^T]^T, 
\end{equation}
\begin{equation}\label{eq:subnet:cnn}
    z_{l,k} = \hat{f}_i^\textrm{CNN}(I_{l,k}|\Theta_i), ~z_{r,k} = \hat{f}_i^\textrm{CNN}(I_{r,k}|\Theta_i),
\end{equation}
\end{subequations}
where $\hat{f}_i^\textrm{CNN}$ denotes the $i$th CNN backbone and embeds the images $I_{l,k}$, $I_{r,k}$ separately into vectors $z_{l,k},z_{r,k}\in\RR^{1024}$ in \eqref{eq:subnet:cnn}; $\hat{f}_i^\textrm{MLP}$ is the $i$th Multi-Layer Perceptron (MLP). It takes the vector $[z_{l,k}^T ~~ z_{r,k}^T]^T$ after the CNN backbone and passes through fully connected layers to generate estimates $\mu_i, \sigma_i^2$ according to \eqref{eq:subnet:mlp:final}, \eqref{eq:subnet:mlp:hidden}; $\Theta_i$ is the CNN parameter; $\{W_i^{(j)}, b_i^{(j)}\}_{j=0}^{\ell_i}$ are the MLP parameters; $\sigma_{\text{ReLU}}$ is an element-wise ReLU activation function; $\ell_i$ is the number of hidden layers in $i$th MLP. Each DNN is trained using the following Negative Log Likelihood (NLL) loss, 
\begin{equation}\label{eq:loss}
    \mathcal{L}\brk{d_k, I_{l,k}, I_{r,k}|\hat{f}_i}  = 
    \log{\sigma_i^2(I_{l,k}, I_{r,k})} + \brk{d_k - \mu_i(I_{l,k}, I_{r,k})}^2 ~/~ \sigma_i^2(I_{l,k}, I_{r,k}).
\end{equation}
This NLL loss implies that if $d_k$ follows a Gaussian distribution $\mathcal{N}\big(\mu_i, \sigma_i^2\big)$, then $\mu_i$ is the estimate of the mean and $\sigma_i^2$ provides a heuristic measure of uncertainty. Inspired by the Deep Ensemble~\citep{deep_ensemble}, we assemble $m$ DNN paths to enable a voting mechanism for outlier detection and generate the final distance headway estimates as a Gaussian mixture as follows,
\begin{equation}\label{eq:mixture}
    [\mu_k~\sigma_k^2]^T = \hat{f}(I_{l,k}, I_{r,k}),
    ~
    \mu_k = \frac{1}{n}\sum_{i=1}^m \mu_i,
    ~
    \sigma^2_k = \frac{1}{m}\sum_{i=1}^m\big(\sigma_i^2 + \mu^2_i\big) - \mu^2_k.
\end{equation}

It should be noted that the above Deep Ensemble provides only a heuristic measure of uncertainty through $\sigma^2_k$, which is an artificial product of the DNN paths, as training using the NLL loss does not require a ground truth label for $\sigma^2_k$. 
We utilize the following results based on Conformal Prediction to define a measure of uncertainty with a statistical guarantee.

\propRmkSpace
\begin{prop}\label{prop:conformal_dk}
    Given a Deep Ensemble $\hat{f}:\I\times\I \to \RR$ pretrained on a training dataset $\D_\textrm{train}$, denoted as $\hat{f}_{\D_\textrm{train}}$, and a calibration dataset $\D_\textrm{cali} = \bcur{I_{l,k}, I_{r,k}, d_k}_{k=1}^n$, where each data point is a triplet comprising images $I_{l,k}$, $I_{r,k}$, and the ground truth distance headway $d_k$, we define the score function,
    $S(I_{l,k}, I_{r,k}, d_k) = \left|\mu_k - d_k\right| / \sigma_k$, where $[\mu_k~\sigma_k^2]^T = \hat{f}_{\D_\textrm{train}}(I_{l,k}, I_{r,k})$. 
    Given a test point $I_{l,\textrm{test}}$, $I_{r,\textrm{test}}$, and the distance headway $d_\textrm{test}$, assuming $\bcur{I_{l,k}, I_{r,k}, d_k}_{k=1}^n$ and $\{ I_{l,\textrm{test}}, I_{r,\textrm{test}}, d_\textrm{test} \}$ are exchangeable, the conformal quantile $q(\alpha)$ from Algorithm~\ref{al:conformal} guarantees that 
    \begin{equation*}
    \prob{d_\textrm{test}\in \C(I_{l,\textrm{test}}, I_{r,\textrm{test}})}\geq 1-\alpha,
    ~
    \C(I_{l,\textrm{test}}, I_{r,\textrm{test}}) = [\mu_\textrm{test}-q(\alpha)\sigma_\textrm{test}, \mu_\textrm{test}+q(\alpha)\sigma_\textrm{test}],
    \end{equation*}    
    where $[\mu_\textrm{test}~\sigma_\textrm{test}^2]^T = \hat{f}_{\D_\textrm{train}}(I_{l,\textrm{test}}, I_{r,\textrm{test}})$.
\end{prop}
\propRmkSpace

\propRmkSpace
\begin{rmk}\label{rmk:conformal_dk}    
Using simulations, we can ensure that data points in $\D_\textrm{cali}$ are collected independently and follow the same distribution by fixing the simulation environment configurations. This satisfies the i.i.d. assumption and also meets the exchangeability assumption, as noted in Remark~\ref{rmk:conformal}.
\end{rmk}
\propRmkSpace

\if\arxiv1
\begin{proof}
Based on Theorem~\ref{thm:conformal} and Remark~\ref{rmk:conformal}, we have 
$
\prob{\left|\mu_\textrm{test} - d_\textrm{test}\right| / \sigma_\textrm{test} \leq q(\alpha)} \geq 1 - \alpha,
$
which implies 
$
\prob{\mu_\textrm{test} - q(\alpha)\sigma_\textrm{test} \leq d_\textrm{test} \leq \mu_\textrm{test} + q(\alpha)\sigma_\textrm{test}} \geq 1 - \alpha.
$
\end{proof}
\fi 

Finally, we can use the results above to provide uncertainty quantification for the measurement model described in \eqref{eq:model:percept} using the following results.
\propRmkSpace
\begin{prop}\label{prop:conformal:box}
    Given an image buffer of $M=2$, i.e., $I_{l,k-1}, I_{r,k-1}, I_{l,k}, I_{r,k}$, and the previous ego vehicle acceleration $a_{k-1}$, we assume that $\D_\textrm{cali}$, $(I_{l,k-1}, I_{r,k-1}, d_{k-1})$ and $\D_\textrm{cali}$, $(I_{l,k}, I_{r,k}, d_k)$ are both exchangeable. Additionally, we assume the ego vehicle's speed $v_k$ is known and the lead vehicle maintains a constant speed between the time instants $t_{k-1}$ and $t_k$. Under these assumptions, the state vector satisfies the following properties:
    \begin{subequations}\label{eq:conformal:box}
    \begin{equation}
        \prob{x_k \in \bsq{x_k}_\alpha } \geq 1 - 2\alpha, 
        ~
        \bsq{x_k}_\alpha := [\Bar{x}_k-q(\alpha) r_k, \Bar{x}_k + q(\alpha) r_k],
    \end{equation}
    \begin{equation}
        \Bar{x}_k = \bsq{\mu_k ~ \mu'_k ~v_k}^T, 
        ~ 
        r_k = \bsq{\sigma_k ~ \sigma'_k ~ 0}^T, 
    \end{equation}
    \begin{equation}\label{eq:conformal_box:Delta_V}        
        \mu'_k = \frac{1}{\Delta t} (\mu_k - \mu_{k-1}) - \frac{1}{2} a_{k-1} \Delta t,
        ~
        \sigma'_k = \frac{1}{\Delta t} (\sigma_k + \sigma_{k-1})
    \end{equation}
    \begin{equation}
        [\mu_k~\sigma_k^2]^T = \hat{f}_{\D_\textrm{train}}(I_{l,k}, I_{r,k}),
        ~
        [\mu_{k-1}~\sigma_{k-1}^2]^T = \hat{f}_{\D_\textrm{train}}(I_{l,k-1}, I_{r,k-1}),
    \end{equation}
    \end{subequations}
    where $\bsq{a,b}\subset \RR^p$ denotes a hypercube with lower bound $a\in \RR^p$ and upper bound $b\in \RR^p$, and $q(\alpha)$ is the conformal quantile from Algorithm~\ref{al:conformal} with the score function $S$ and the pre-trained Deep Ensemble $\hat{f}_{\D_\textrm{train}}$ as defined in Proposition~\ref{prop:conformal_dk}.
\end{prop}
\propRmkSpace

\if\arxiv1
\begin{proof}
    We define the following probabilistic events in the probability space $(\X, \P(\X), \PP)$,
    \begin{equation*}
        E_k = \bcur{ d_k \in [\mu_k-q(\alpha) \sigma_k, \mu_k + q(\alpha) \mu_k] },
        ~ \text{and} ~
        E'_k = \bcur{ \Delta v_k \in [\mu'_k-q(\alpha) \sigma'_k, \mu'_k + q(\alpha) \mu'_k] },
    \end{equation*}
    where $x_k\in \X\subset \RR^3$, $\P(\X)$ is the power set of $\X$, and $\PP$ is the probability measure. Then, we can derive the following equations,
    \begin{equation*}
    \begin{aligned}     
        & \prob{x_k \in \bsq{x_k}_\alpha | a_{k-1}, v_k, \D_\textrm{cali}, I_{l,k-1}, I_{r,k-1}, I_{l,k}, I_{r,k}} 
        \\
        & = \prob{E_k, E'_k,v_k | a_{k-1}, v_k, \D_\textrm{cali}, I_{l,k-1}, I_{r,k-1}, I_{l,k}, I_{r,k}} 
        \\
        & = \prob{E_k, E'_k,v_k | a_{k-1}, v_k, E_k, E_{k-1}} \cdot \prob{E_k, E_{k-1} | \D_\textrm{cali}, I_{l,k-1}, I_{r,k-1}, I_{l,k}, I_{r,k}} 
        \\
        & = \prob{E'_k | a_{k-1}, v_k, E_k, E_{k-1}} \cdot \prob{E_k, E_{k-1} | \D_\textrm{cali}, I_{l,k-1}, I_{r,k-1}, I_{l,k}, I_{r,k}}  
        \\
        & = 1 \cdot \prob{E_k, E_{k-1} | \D_\textrm{cali}, I_{l,k-1}, I_{r,k-1}, I_{l,k}, I_{r,k}}
        \\
        & = 1 - \prob{E_k^c \cup E^c_{k-1} | \D_\textrm{cali}, I_{l,k-1}, I_{r,k-1}, I_{l,k}, I_{r,k}}
        \\
        & \geq 1 - \Brk{ \prob{E_k^c | \D_\textrm{cali}, I_{l,k}, I_{r,k}} + \prob{ E^c_{k-1} | \D_\textrm{cali}, I_{l,k-1}, I_{r,k-1}} } 
        \\
        & = 1 - \Brk{ 1-  \prob{E_k | \D_\textrm{cali}, I_{l,k}, I_{r,k}} + 1 - \prob{ E_{k-1} | \D_\textrm{cali}, I_{l,k-1}, I_{r,k-1}} }
        \\
        & =  \prob{E_k | \D_\textrm{cali}, I_{l,k}, I_{r,k}} + \prob{ E_{k-1} | \D_\textrm{cali}, I_{l,k-1}, I_{r,k-1}} - 1 
        \\
        & \geq (1-\alpha) +  (1-\alpha) - 1 = 1-2\alpha
        , 
    \end{aligned}
    \end{equation*}
    where the second equality follows from the marginalization; We note that $\Delta v = \frac{1}{\Delta t} (d_k - d_{k-1}) - \frac{1}{2} a_{k-1} \Delta t$ due to the constant lead vehicle speed assumption followed from \eqref{eq:model:kine}. The third equality is due to $E'_k$ being true given $a_{k-1}, v_k, E_k, E_{k-1}$, and the construction in \eqref{eq:conformal_box:Delta_V}; The fifth equality is derived from De Morgan's Law where $E^c_{k-1}$ and $E^c_{k}$ are the complement sets, i.e., $E^c_{k} = \X \backslash E_k$ and $E^c_{k-1} = \X \backslash E_{k-1}$; The first inequality follows from Boole's inequality; The final inequality is derived from Conformal Prediction, given the exchangeability assumptions.
\end{proof}
\fi

% ==================================================
\subsection{Conformal Tube Model Predictive Control}\label{subsec:method:control} 

At the current time $t_k$, we assume the following quantities are given: previous acceleration $a_{k-1}$, an image buffer of $M=2$, i.e., $I_{l,k-1}, I_{r,k-1}, I_{l,k}, I_{r,k}$, and the vehicle speed $v_k$. Note that the actual distance headway, i.e., $d_{k-1}$ and $d_k$, is unknown to the algorithm. We can then predict the trajectory of future distance headway as a tube formed by the Conformal Prediction sets for a variable acceleration trajectory using the following proposition:

\propRmkSpace
\begin{prop}\label{prop:conformal:tube}
    Given the conformal quantile $q(\alpha)$, the score function $S$, and the pre-trained Deep Ensemble $\hat{f}_{\D_\textrm{train}}$ in Proposition~\ref{prop:conformal:box}, for a variable acceleration sequence $\{a_{k+i}\}_{i=0}^{N-1}$, the prediction tube $\{\bsq{x_{k+i}}_\alpha\}_{i=0}^{N}$ formed by Conformal Prediction sets possesses the following properties:
    \begin{subequations}\label{eq:conformal:tube}
    \begin{equation}\label{eq:conformal:tube:prob}
        \prob{ \cap_{i=0}^{N}{\bcur{x_{k+i} \in \bsq{x_{k+i}}_\alpha}} } \geq 1 - 2\alpha, 
    \end{equation}
    \begin{equation}\label{eq:conformal:tube:box}
        \bsq{x_{k+i}}_\alpha := \bsq{\Bar{x}_{k+i} - q(\alpha)r_{k+i}, ~\Bar{x}_{k+i} + q(\alpha)r_{k+i}},
    \end{equation}
    \begin{equation}\label{eq:conformal:tube:centriod_size}
        \Bar{x}_{k+i} = A\Bar{x}_{k+i-1} + B a_{k+i-1},
        ~
        r_{k+i} = A^\text{abs} r_{k+i-1},
        ~
        i = 1,\dots, N,
    \end{equation}
    \end{subequations}
    where $\Bar{x}_k$ and $r_k$ are defined in Proposition~\ref{prop:conformal:box}, and $A^\text{abs}$ represents the element-wise absolute operation on matrix $A$. The prediction tube $\bcur{\bsq{x_{k+i}}_\alpha}_{i=0}^{N}$, with trajectories of centroids $\brk{\Bar{x}_{k+i}}_{i=0}^{N}$ and half-sizes $\brk{r_{k+i}}_{i=0}^{N}$, can achieve a coverage with a probabilistic lower bound of $(1-2\alpha)$.
\end{prop}
\propRmkSpace

\if\arxiv1
\begin{proof}
    We define probabilistic events $F_{k+i} =\bcur{ x_{k+1}\in\bsq{x_{k+1}}_{\alpha}} \subset \X$, $i=0,\dots,N$, in probability space $(\X, \P(\X), \PP)$. By construction in \eqref{eq:conformal:tube}, we have $\prob{F_{k+i} | F_{k+i-1}, a_{k+i-1}} = 1$ following from the kinematics \eqref{eq:model:kine}. Then, the following equalities and inequality hold,
    \begin{equation*}
    \begin{array}{c}
       \prob{ \cap_{i=0}^{N}F_{k+i} | \{a_{k+i}\}_{i=0}^{N-1}} 
        = \prob{F_{k+N} | F_{k+N-1}, a_{k+N-1}}
        \cdot
        \prob{F_{k+N-1} | F_{k+N-2}, a_{k+N-2}}
        \\
        \cdots       
        \prob{F_{k+1} | F_{k}, a_{k}} 
        \cdot\prob{F_k}\cdot\prob{a_k} 
        = \prob{F_k} \geq 1 - 2\alpha,
    \end{array}
    \end{equation*}
    where the last inequality is the conformal coverage guarantee provided by  Proposition~\ref{prop:conformal:box}.
\end{proof}
\fi

Hence, treating $(a_{k+i})_{i=0}^{N-1}$ as decision variables, we formulate an MPC problem that predicts future trajectories as a Conformal Tube $\{\bsq{x_{k+i}}_\alpha\}_{i=0}^{N}$ for different $(a_{k+i})_{i=0}^{N-1}$ and optimizes $(a_{k+i})_{i=0}^{N-1}$ while incorporating the objectives in Sec.~\ref{sec:problem} according to
\begin{subequations}\label{eq:MPC}
    \begin{multline}\label{eq:MPC:obj}
        \argmin\limits_{\substack{a_{k+i-1}, \Bar{x}_{k+i}, r_{k+i},\\ i=1,\dots,N}}
        ~
        \sum_{i=0}^{N-1} r_1 a^2_{k+i} + r_2 (a_{k+i}-a_{k+i-1})^2 
         + \sum_{i=1}^N  (\Bar{x}_{k+i} - x_s)^T Q (\Bar{x}_{k+i} - x_s)
    \end{multline}
    \begin{equation} \label{eq:MPC:safe}
        \text{ subject to: \hspace{1cm}}
        \eqref{eq:conformal:tube:box}, 
        ~
        \eqref{eq:conformal:tube:centriod_size},
        \hspace{1cm}
        Cx_{k+i} \leq b, ~\forall x_{k+i}\in \bsq{x_{k+i}}_\alpha
    \end{equation}
    \begin{equation}\label{eq:MPC:acc}
        a_{\min}\leq a_{k+i-1}\leq a_{\max},~ i= 1,\dots,N,
    \end{equation}
\end{subequations}
where $ x_s = [0~0~v_s]^T $ represents the desired state, $ v_s $ is the user-defined target speed, $Q = \text{diag}([0~q_1~q_2]^T) $ is weight matrix,  and the weights $ r_1, r_2, q_1, q_2 > 0 $ are used to balance the minimization of control effort, the reduction of the rate of changes in control, lead vehicle-following, and speed tracking, respectively. Note that acceleration limits are enforced through the constraint \eqref{eq:MPC:acc}. The constraints \eqref{eq:conformal:tube:box}, \eqref{eq:conformal:tube:centriod_size}, \eqref{eq:MPC:safe} ensure a safe following distance and impose speed limits on the feasible states in the Conformal Prediction Tube with
\begin{equation*}
    C = 
    \scalebox{0.8}{$
    \begin{bmatrix}
        -1 & 0 & T_s\\
        0 & 0 & 1\\
        0 & 0 & -1\\
    \end{bmatrix}
    $}
    ,~
    b = 
    \scalebox{0.8}{$
    \begin{bmatrix}
        - d_s\\ v_\text{max} \\ - v_\text{min}
    \end{bmatrix}
    $}
    .
\end{equation*}
Here, $d_s$, $T_s$ are the adjustable ACC stopping distance, and constant time headway, respectively. The solution $(a_{k+i})_{i=0}^{N-1}$ of the MPC ~\eqref{eq:MPC}, if it exists, provides a probabilistic safety guarantee, as demonstrated by the following results:
\propRmkSpace
\begin{prop}\label{prop:MPC}
    Given the quantile $q(\alpha)$, the score function $S$, and the pre-trained Deep Ensemble $\hat{f}_{\D_\textrm{train}}$ in Proposition~\ref{prop:conformal:tube}, the solution $\{a_{k+i}\}_{i=0}^{N-1}$ of MPC \eqref{eq:MPC}, if it exists, ensures that
    \begin{equation}
        \prob{ \cap_{i=0}^{N}\bcur{ Cx_{k+i} \leq b} } \geq 1 - 2\alpha,
    \end{equation}
    which implies that the probability of satisfying the safety constraints over the prediction horizon of length $N$ is at least $(1 - 2\alpha)$.
\end{prop}
\propRmkSpace

\if\arxiv1
\begin{proof}  
    The solution $\{a_{k+i}\}_{i=0}^{N-1}$ yields a Conformal Prediction tube $\{\bsq{x_{k+i}}_\alpha\}_{i=0}^{N}$. According to Propostiion~\ref{prop:conformal:tube} and the construction of constraints \eqref{eq:MPC:safe}, the following equalities can be derived,
    \begin{equation*}
        \prob{ \cap_{i=0}^{N}\bcur{ Cx_{k+i} \leq b} | \{a_{k+i}\}_{i=0}^{N-1}}
        = \prob{ \cap_{i=0}^{N}F_{k+i} | \{a_{k+i}\}_{i=0}^{N-1}} \geq 1 - 2\alpha.
    \end{equation*}
\end{proof}
\fi

It is important to note that the constraints \eqref{eq:MPC:safe} may not always be feasible, depending on the selected miscoverage rate $\alpha$ and the associated conformal quantile $q(\alpha)$. A lower miscoverage rate $\alpha$ corresponds to a larger conformal quantile $q(\alpha)$, resulting in a more relaxed Conformal Prediction Tube as described in Proposition~\ref{prop:conformal:tube}, thus tightening the constraints \eqref{eq:MPC:safe}. To address this issue, we develop the following results, which ensure the recursive feasibility of the Conformal Tube MPC \eqref{eq:MPC}, while continuously monitoring a lower bound of the probability of safety:
\propRmkSpace
\begin{prop}\label{prop:QP_MPC}
    Given the score function $S$, calibration dataset $\D_\textrm{cali}$, and the pre-trained Deep Ensemble $\hat{f}_{\D_\textrm{train}}$ in Proposition~\ref{prop:conformal:tube}, we define the following \textbf{Quadratic Program (QP)},
    \begin{subequations}\label{eq:QP_MPC}
    \begin{equation}\label{eq:QP_MPC:obj}
        \argmin\limits_{\substack{a_{k+i-1}, \Bar{x}_{k+i}, r_{k+i},\\ i=1,\dots,N, ~ \hat{q}}}
        ~
        - \rho \hat{q} + 
        \scalebox{0.88}{
        $
        \sum_{i=0}^{N-1} r_1 a^2_{k+i} + r_2 (a_{k+i}-a_{k+i-1})^2 
         + \sum_{i=1}^N  (\Bar{x}_{k+i} - x_s)^T Q (\Bar{x}_{k+i} - x_s)
         $
         }
    \end{equation}
    \begin{equation}\label{eq:QP_MPC:safe}
        \text{ subject to: }
        ~
        C\Bar{x}_{k+i} + \hat{q} \cdot C^\text{abs} r_{k+i} \leq b,
        ~    
        \Bar{x}_{k+i} = A\Bar{x}_{k+i-1} + B a_{k+i-1},
        ~
        r_{k+i} = A^\text{abs} r_{k+i-1},  
    \end{equation}
    \begin{equation}\label{eq:QP_MPC:acc}    
        a_{\min}\leq a_{k+i-1}\leq a_{\max},~ i= 1,\dots,N,
    \end{equation}
    \end{subequations}
    where $\hat{q}\in\RR$ is a new decision variable, $\rho>0$ is a tunable parameter for penalizing small $\hat{q}$, and $\Bar{x}_k$ and $r_k$ are defined according to Proposition~\ref{prop:conformal:box}. Then, the solution $(a_{k+i})_{i=0}^{N-1}$, $\hat{q}$, which always exists, possesses the following properties:
    \begin{subequations}\label{eq:QP_MPC:tube}
    \begin{equation}\label{eq:QP_MPC:safe:prob}
        \prob{ 
        \cap_{i=0}^{N}\bcur{ 
        Cx_{k+i} \leq b
        } 
        } \geq 1 - 2\hat{\alpha}, 
    \end{equation}
    \begin{equation}\label{eq:QP_MPC:safe:alpha}
        \hat{\alpha} = 1 - \frac{\hat{n}}{n+1},
        ~ 
        \hat{n} =  \abs{ \bcur{(I_{l,i}, I_{r,i}, d_i) \in\D_\textrm{cali}:~  S(I_{l,i}, I_{r,i}, d_i)\leq \hat{q}} } .
    \end{equation}
    \end{subequations}
\end{prop}
\propRmkSpace

\if\arxiv1
\begin{proof}    
    By construction in \eqref{eq:QP_MPC:safe:alpha}, we have $\ceil{(n+1)(1-\hat{\alpha})}=\hat{n}$ which implies $q = \hat{q}$ for this particular $\alpha=\hat{\alpha}$ based on Algorithm~\ref{al:conformal}. Then, the solution  $\{a_{k+i-1}, \Bar{x}_{k+i}, r_{k+i}, i=1, \dots, N, ~ \hat{q}\}$ constructs a tube $\bcur{\bsq{x_{k+i}}_{\hat{\alpha}}}_{i=0}^{N}$ and
    $\bsq{x_{k+i}}_{\hat{\alpha}} = \bsq{\Bar{x}_{k+i} - \hat{q} r_{k+i}, ~\Bar{x}_{k+i} + \hat{q} r_{k+i}}$, which provides a probabilistic lower bound of safety similar to Proposition~\ref{prop:MPC}.
\end{proof}
\fi

\propRmkSpace
\begin{rmk}
    The solutions $\hat{q}$ and $(a_{k+i})_{i=0}^{N-1}$ inform a Conformal Prediction Tube $\bcur{\bsq{x_{k+i}}_{\hat{\alpha}}}_{i=0}^{N}$, where $\bsq{x_{k+i}}_{\hat{\alpha}} = \bsq{\Bar{x}_{k+i} - \hat{q} r_{k+i}, ~ \Bar{x}_{k+i} + \hat{q} r_{k+i}}$ according to \eqref{eq:QP_MPC:safe}. 
    The equations \eqref{eq:QP_MPC:safe:alpha}, \eqref{eq:QP_MPC:safe:prob} imply that, 
    within this tube, the state trajectory is maintained, thereby ensuring that the probability of satisfying the safety constraints across $N$ prediction horizons is at least $(1 - 2\hat{\alpha})$.
    Unlike the typical setting, $\hat{\alpha}$ depends on the decision variable $\hat{q}$. The optimization problem \eqref{eq:QP_MPC} maximizes $\hat{q}$, which subsequently minimizes the miscoverage rate $\hat{\alpha}$ and maximizes the probabilistic lower bound of safety, i.e., $(1 - 2\hat{\alpha})$.
\end{rmk}
\propRmkSpace

\propRmkSpace
\begin{rmk}
    We also note that the Conformal Prediction Tube is empty when the solution $\hat{q} < 0$, corresponding to situations where no acceleration trajectories can safely regulate the system with respect to the constraints \eqref{eq:MPC:safe}. Such scenarios are related to contingency planning. In these cases, we set $a_k = a_\text{min}$, which serves as an effective strategy in emergencies.
\end{rmk}
\propRmkSpace

% ==================================================
\section{Results}\label{sec:result}
We train the Deep Ensemble and apply neuron pruning to improve memory usage (Sec.~\ref{subsec:results:training}). We then present distance headway estimation results using the Deep Ensemble with Conformal Prediction, demonstrating reliable uncertainty characterization in various scenarios, including OOD and adversarial attacks (Sec.~\ref{subsec:results:perception}). Finally, we evaluate the Conformal Tube MPC design in ACC scenarios using a realistic simulator (Sec.~\ref{subsec:results:control}). Computations are performed on an Nvidia GeForce RTX 4080 GPU (16 GB) and a 13th Gen Intel i9-13900F CPU (32 GB RAM).

% ==================================================
\subsection{Deep Ensemble Training and Pruning}\label{subsec:results:training} 

\begin{figure}[!h]
\begin{center}
\if\arxiv0
\includegraphics[width=0.9\linewidth]{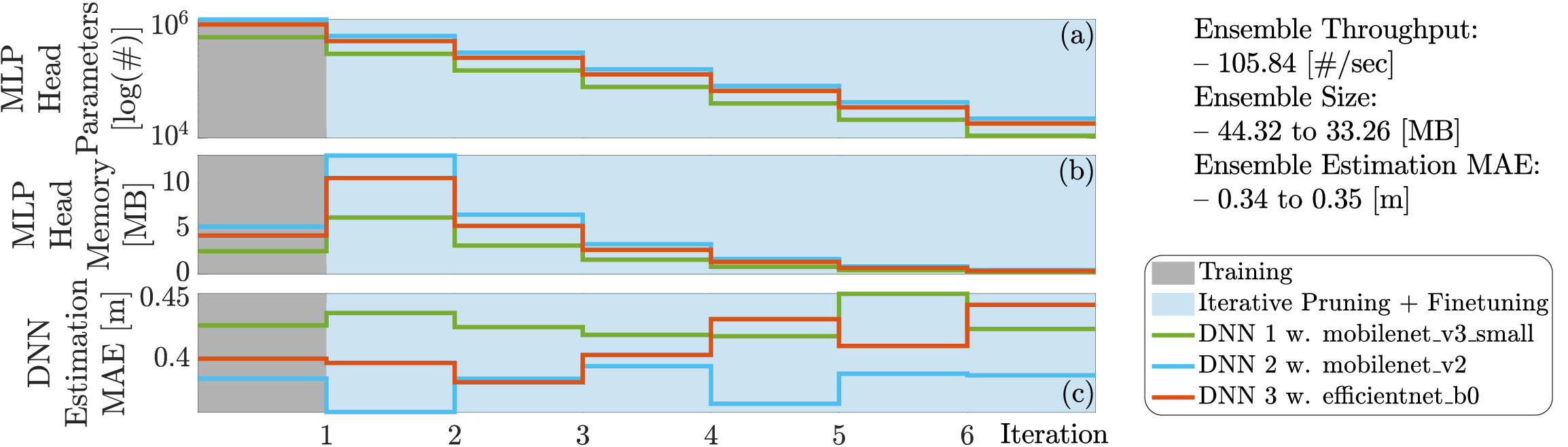}
\fi
\if\arxiv1
\includegraphics[width=0.9\linewidth]{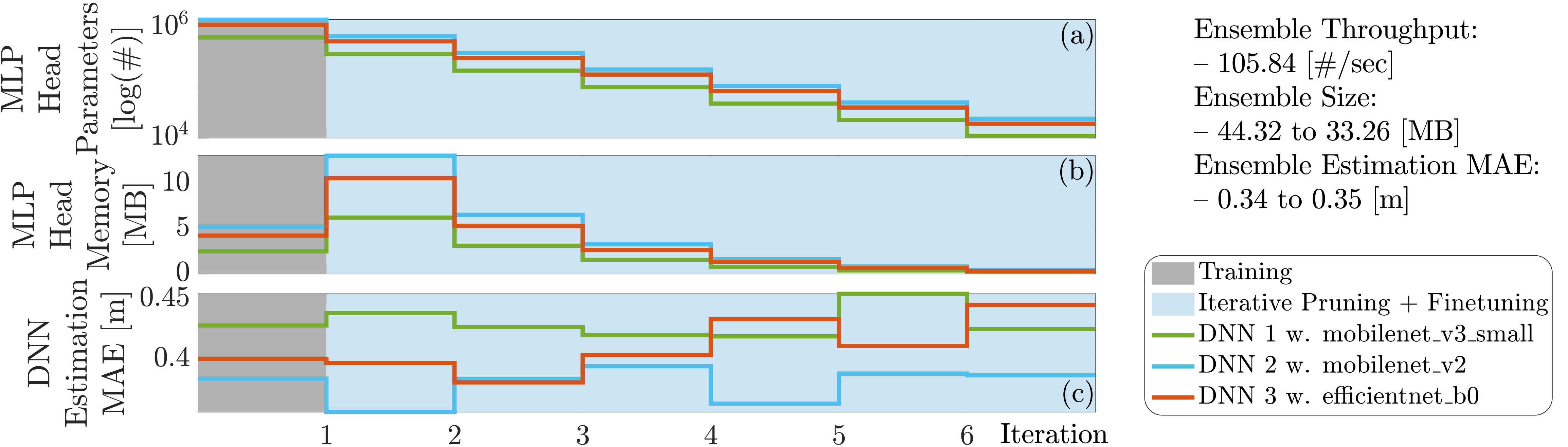}
\fi
\end{center}
\caption{Memory and MAE statistics of three DNN paths during the iterative pruning process: (a) Number of parameters in each DNN path.
(b) Memory required for parameter storage in megabytes (MB). (c) MAE after initial training and after each iteration of fine-tuning post-pruning.}
\label{fig:pruning}
\end{figure}

% =========================================================
\if\arxiv1
In the Deep Ensemble, we implement three DNN paths using MobileNet V2~\citep{mobilenet_v2}, MobileNet V3~\citep{mobilenet_v3}, and EfficientNet~\citep{efficientnet} as CNN backbones, respectively, utilizing PyTorch~\citep{pytorch}. These networks are selected due to their outstanding performance in image classification tasks and lower computational demands. 
Meanwhile, we use the same MLP head architecture for the three DNN paths, each having $\ell_{i=1,2,3}=2$ hidden layers, with 512 and 128 neurons, respectively.
For dataset collection and testing, we utilize the Carla simulator~\citep{carla}. We generate a dataset $\mathcal{D}_\text{train} = \bcur{d_k, I_{l,k}, I_{r,k}}_{k=1}^{29580}$ consisting of 29,580 data triplets. Data points are collected within the map \texttt{Town06} in Carla. To simplify the approach exposition, we fix the model of the lead vehicle to a 2020 Lincoln MKZ Sedan (\texttt{vehicle.lincoln.mkz\_2020}) and set the weather to \texttt{ClearNoon} (good lighting conditions, no rain, and no objects casting shadows). 
We train the three DNN paths separately using Stochastic Batch Gradient Descent with a learning rate of 0.001, a momentum of 0.9, and batch sizes of 65, 65, and 60, respectively, for 100 epochs. To ensure numerical stability with the logarithm in the loss function \eqref{eq:loss}, we enforce the positiveness of the output variance $\sigma_i^2$ using the following assignment, $\sigma_i^2 \leftarrow \epsilon + \log\brk{1 + \exp{\sigma_i^2}}$, where a small $\epsilon = 10^{-6}$ is chosen.

To further reduce the size of the DNN paths, we iteratively prune the parameters $\{W_i^{(j)}\}_{j=0}^{\ell_i}$ in the dense MLP heads based on their magnitude. In each iteration, we first trim 50\% of the remaining weight parameters in each of the three MLP heads, then finetune for 5 epochs using the same settings as in the initial training. 
This pruning and fine-tuning process is repeated for 6 iterations, resulting in a Deep Ensemble with memory usage reduced from 44.32 MB to 33.26 MB. As shown in Fig.~\ref{fig:pruning}, despite the significant reduction in memory usage, the Mean Absolute Error (MAE) of the Deep Ensemble estimation, i.e., $\E \abs{\mu_k - d_k}$, only increases slightly by 0.01 meters. 
We also note that memory usage increases during the first iteration, as shown in Fig.~\ref{fig:pruning}b. This increase is due to storing the pruned parameters using sparse matrices, where each weight occupies 4 bytes and each location index requires two integers, each taking 8 bytes.
\fi

% =========================================================
\if\arxiv0
Using PyTorch~\citep{pytorch}, the Deep Ensemble leverages three DNN paths with MobileNet V2~\citep{mobilenet_v2}, MobileNet V3~\citep{mobilenet_v3}, and EfficientNet~\citep{efficientnet} as CNN backbones, chosen for their strong image classification performance and efficiency. Dataset creation and testing occur in the Carla simulator~\citep{carla} in the \texttt{Town06} map. The lead vehicle model is chosen as a 2020 Lincoln MKZ Sedan, and the weather is set to \texttt{ClearNoon} for consistency. To reduce the size of the DNN paths, post-training iterative pruning is applied to the dense MLP heads, removing 50\% of the lowest-magnitude weights in each iteration followed by fine-tuning for 5 epochs. After 6 iterations, memory usage decreases from 44.32 MB to 33.26 MB, with only a minimal MAE increase of 0.01 meters in the ensemble's predictions. Initial memory usage rises due to sparse matrix storage overhead, where each weight and its location index require additional memory. We refer to \fullpaper~for details on DNN design, training, and pruning. 
\fi

\begin{rmk}
    Instead of starting with smaller MLPs, it is generally more effective to begin with larger MLPs and then prune and fine-tune them. Larger networks have greater representational capacity and can capture more complex patterns, resulting in better initial performance. This discussion is often referred to as the ``lottery ticket hypothesis"~\citep{frankle2018lottery}.
\end{rmk}

% ==================================================
\subsection{Distance Headway Estimation}\label{subsec:results:perception} 

\begin{figure}[!h]
\begin{center}
\if\arxiv0
\includegraphics[width=0.95\linewidth]{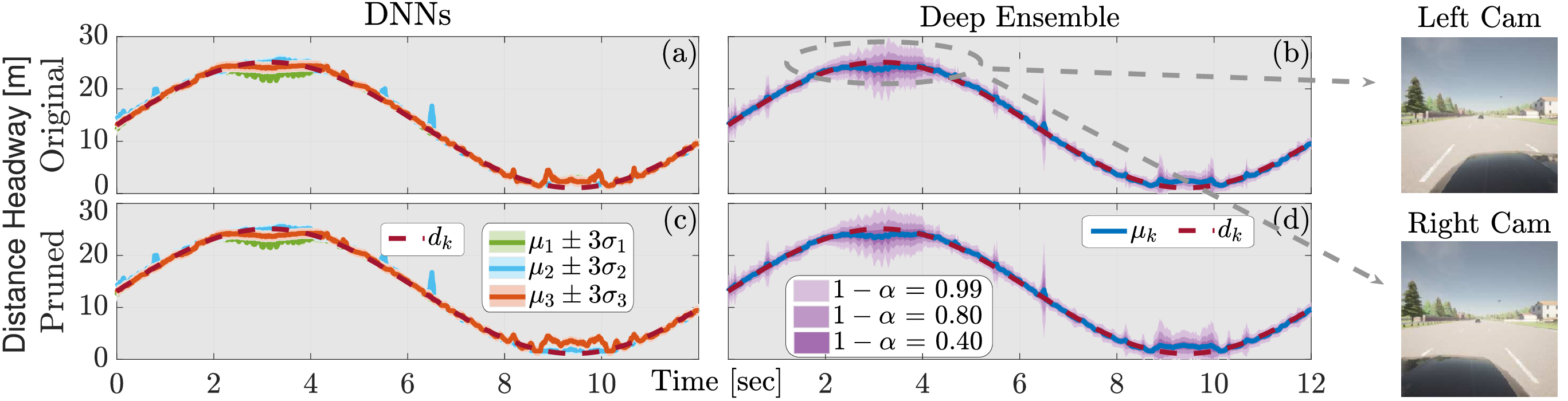}
\fi
\if\arxiv1
\includegraphics[width=0.95\linewidth]{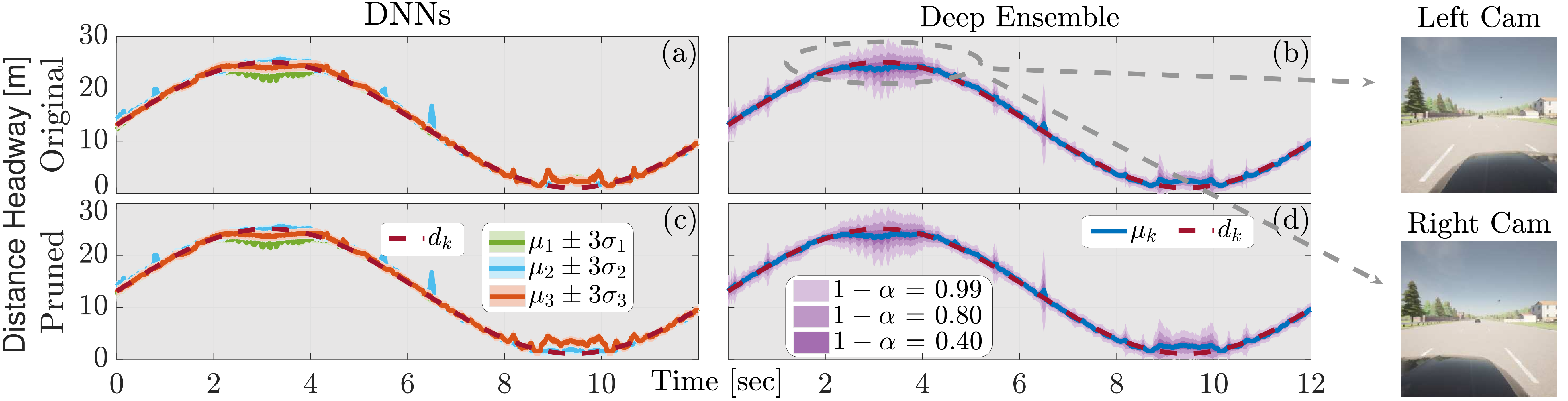}
\fi
\end{center}
\caption{Distance headway estimation results with the ground truth \(d_k\) trajectory shown in red: Mean $\mu_i$ and variance $\sigma_i^2$ estimations from the three DNNs $(i=1, 2, 3)$ are visualized with colored solid lines and bands. Panels (a) and (c) display results for pre-pruning and post-pruning DNNs, respectively. The ensemble results are visualized in panels (b) and (d), with the Conformal Prediction sets $\C(I_{l,k}, I_{r,k})$ of different coverage rates $(1-\alpha)$ plotted as purple bands along the trajectory $\mu_k$. }
\label{fig:ensemble}
\end{figure}

\begin{figure}[!h]
\begin{center}
\if\arxiv0
\includegraphics[width=0.99\linewidth]{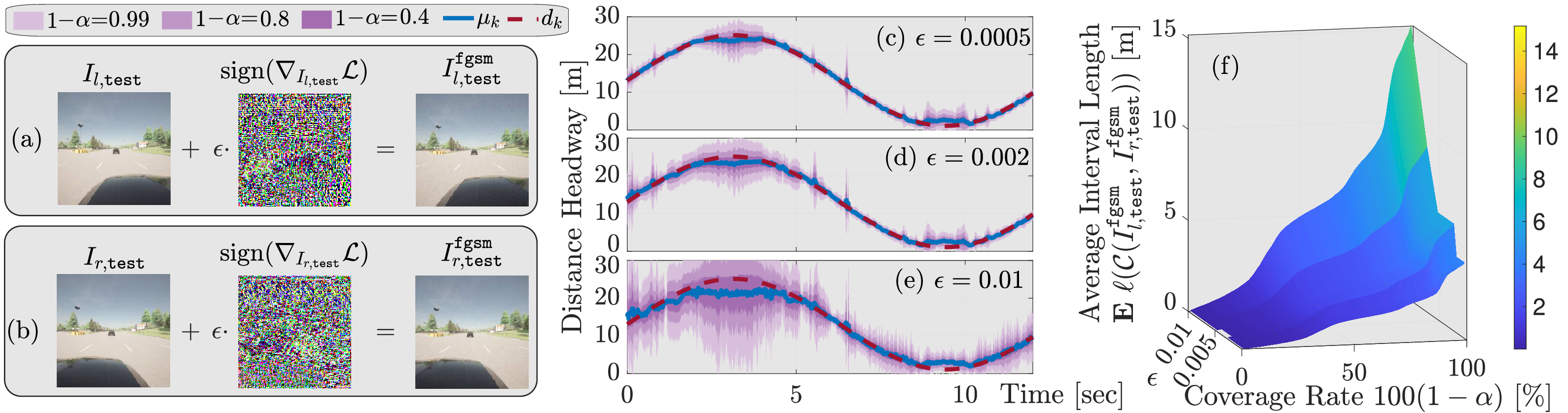}
\fi
\if\arxiv1
\includegraphics[width=0.99\linewidth]{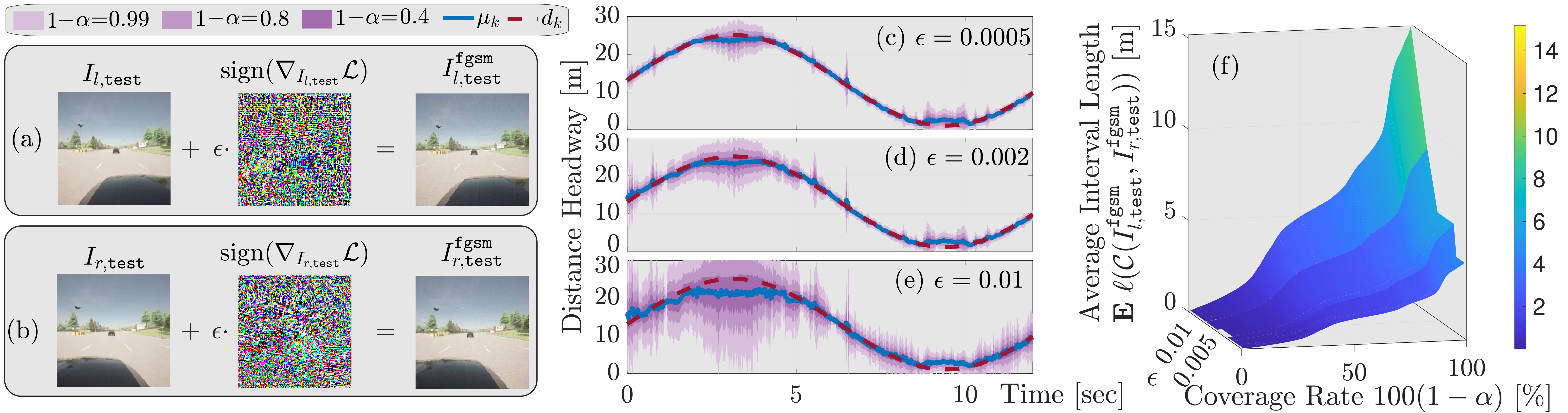}
\fi
\end{center}
\caption{Uncertainty quantification using Deep Ensemble against FGSM adversarial attacks: (a) and (b) show examples of FGSM attackers on the left and right images, respectively.
(c), (d), and (e) depict Conformal Prediction sets $\mathcal{C}(I_{l,k}, I_{r,k})$ with different coverage rates $(1-\alpha)$ under FGSM attacks of varying magnitudes $\epsilon$.
(f) shows the size of the Conformal Prediction sets required to achieve a certain coverage rate against different magnitudes of the attack $\epsilon$. }
\label{fig:adversarial}
\end{figure}

\begin{figure}[!h]
\begin{center}
\if\arxiv0
\includegraphics[width=0.9\linewidth]{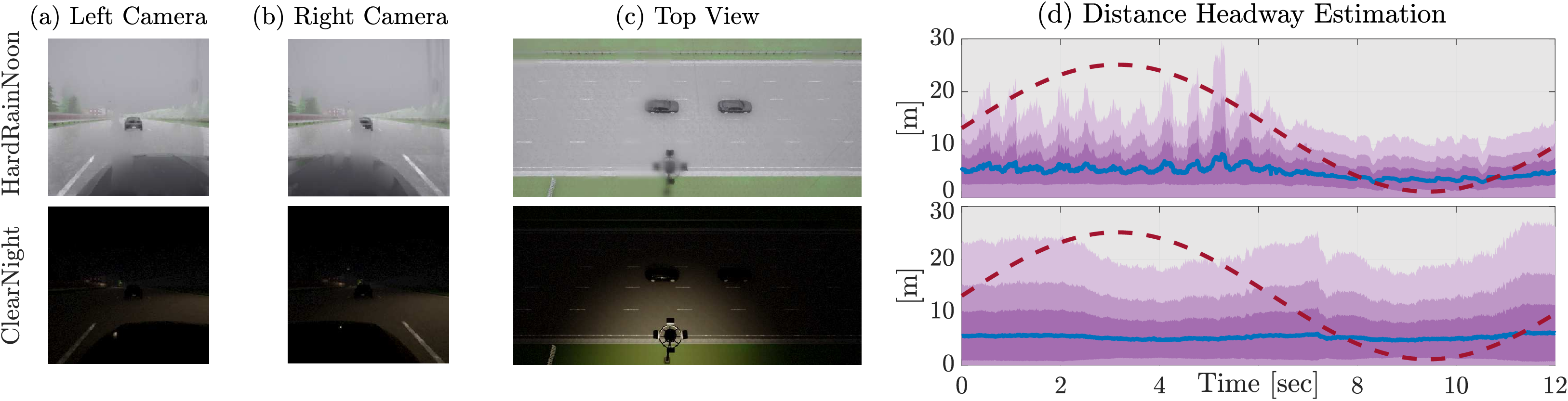}
\fi
\if\arxiv1
\includegraphics[width=0.9\linewidth]{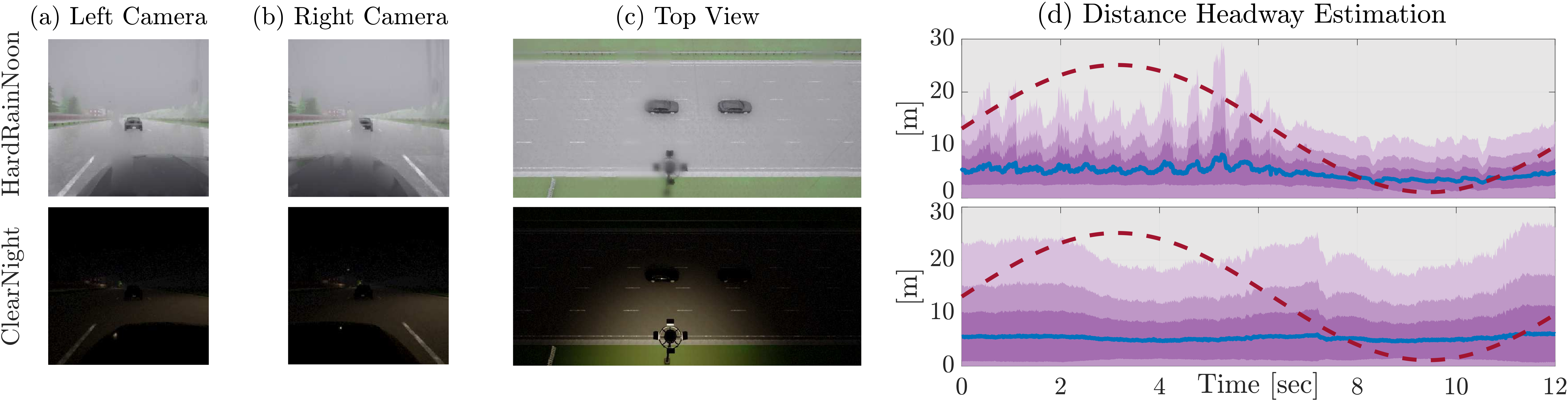}
\fi
\end{center}
\caption{Uncertainty quantification using Deep Ensemble under OOD scenarios induced by different weather conditions is examined. The {\tt ClearNoon} weather setting is used for training, while \texttt{HardRainNoon} and \texttt{ClearNight} are categorized as OOD scenarios.}
\label{fig:ood}
\end{figure}

As shown in Fig~\ref{fig:ensemble}, we collected a testing dataset $\mathcal{D}_\textrm{test}$ for a varying $d_k$ trajectory following the same settings as in training to emulate an in-distribution scenario.
The Conformal Prediction is conducted using a separately collected calibration dataset $\mathcal{D}_\textrm{cali} = \bcur{d_k, I_{l,k}, I_{r,k}}_{k=1}^{10000}$. Our Deep Ensemble, combined with the Conformal Prediction sets $\mathcal{C}(I_{l,k}, I_{r,k})$, provides good coverage of the actual trajectory of $d_k$ with a coverage rate $(1-\alpha)=0.8$, which empirically validates Proposition~\ref{prop:conformal_dk}.
Due to limitations in image resolution, the lead vehicle appears as a black pixel in the RGB images when $d_k \geq 20$ m (see Fig~\ref{fig:ensemble}). We observe that the Deep Ensemble captures this source of uncertainty by presenting larger prediction sets in such cases.

Meanwhile, we perturb the images in the testing datasets using the Fast Gradient Sign Method (FGSM)~\citep{goodfellow2016deep} to examine the uncertainty detection ability of the proposed approach when facing adversarial attacks at the pixel level. 
Specifically, the FGSM perturbs the images along the direction of $\nabla \mathcal{L}_{I_{l,\text{test}}}$ and $\nabla \mathcal{L}_{I_{r,\text{test}}}$ in the image space $\mathcal{I}$, corresponding to the direction where the loss function increases most rapidly.
As shown in Fig.~\ref{fig:adversarial}c-e, our method can reflect the magnitude of the perturbation $\epsilon$ through the changes in the length of the Conformal Prediction sets/intervals, indicated by the purple bands. 
Meanwhile, the length of the Conformal Prediction sets/intervals required to achieve a certain probability of coverage increases with larger perturbations (see Fig.~\ref{fig:adversarial}f).
Lastly, we also manipulate the weather in the Carla simulator (see Fig.~\ref{fig:ood}) to create artificial OOD scenarios. Our method detects OOD by producing larger Conformal Prediction sets, thereby informing the downstream Conformal Tube MPC of the perception uncertainties.

% ==================================================
\subsection{Adaptive Cruise Control}\label{subsec:results:control} 

\begin{figure}[!h]
\begin{center}
\if\arxiv0
\includegraphics[width=0.99\linewidth]{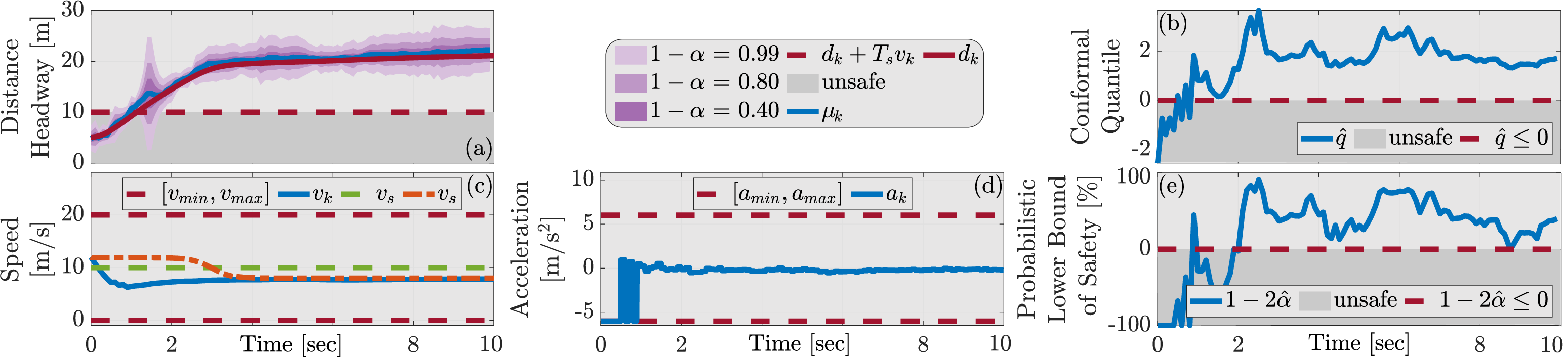}
\fi
\if\arxiv1
\includegraphics[width=0.99\linewidth]{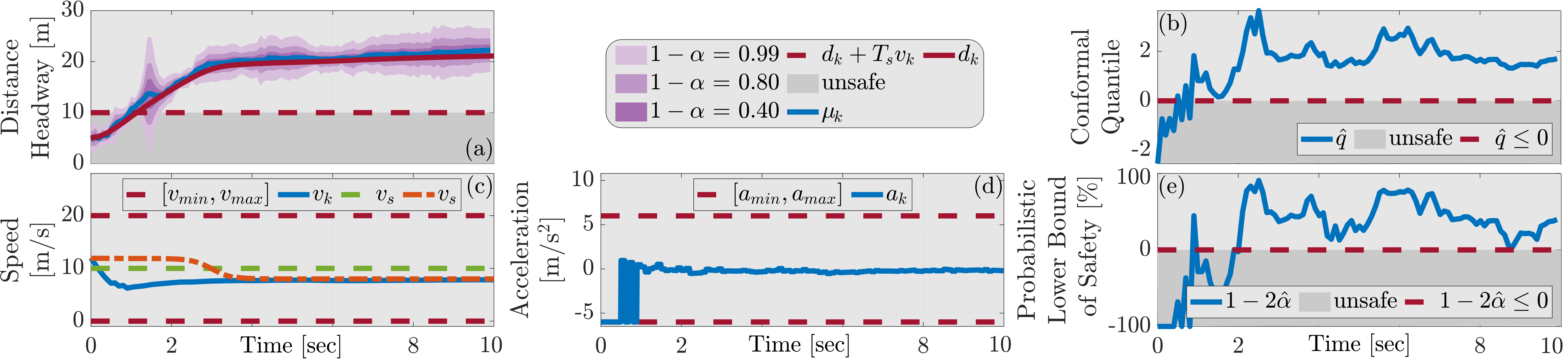}
\fi
\end{center}
\caption{ACC experiment: (a) Distance headway estimation. (b) Trajectory of the solved conformal quantile from QP \eqref{eq:QP_MPC}. (c) Speed trajectories of the ego and the lead vehicle. (d) Acceleration profile. (e) Results showing the probabilistic lower bound of safety from Proposition~\ref{prop:QP_MPC}.}
\label{fig:ctrl_ind}
\end{figure}

We use the Carla simulator to test the Conformal Tube MPC \eqref{eq:QP_MPC} in realistic ACC scenarios. The simulation parameters are set as follows: $[v_{\min}, v_{\max}]=[0,20]\;\rm m/s$ and $[a_{\min},a_{\max}]=[-6, 6]\;\rm m/s^2$.
The Conformal Tube MPC has a prediction horizon of 3 sec, i.e., $N=3$ and $\Delta t=1$ sec. Furthermore, the MPC operates in an asynchronous updating scheme and recomputes $a_k$ every $0.1$ sec. 
Other parameters are set using the following values: $d_s=10$ m, $T_s=0$ sec, and $[r_1,r_2,q_1,q_2,\rho]=[1,5,1,10,100]$. The Conformal Tube MPC problem~\eqref{eq:QP_MPC} is solved using PyDrake~\citep{drake}. 
The perception using Deep Ensemble and Conformal Prediction takes $1.3699\times 10^{-2}\pm 0.3686\times 10^{-2}~\textrm{sec}$ while the solving the QP \eqref{eq:QP_MPC} takes $6.2435\times 10^{-5}\pm 0.6222 \times 10^{-5}~\textrm{sec}$. 
% $30\sim100~\textrm{Hz}$
%
As shown in Fig.~\ref{fig:ctrl_ind}, the ego vehicle initially maintains an unsafe distance headway and consequently applies hard brakes. 
After 2 seconds, upon returning to a safe zone, the lead vehicle decelerates, and the ego vehicle, rather than tracking the set speed \(v_s\), adjusts its speed to follow that of the lead vehicle (see Fig.~\ref{fig:ctrl_ind}c), while maintaining a safe headway distance (see Fig.~\ref{fig:ctrl_ind}a). The safety is formally guaranteed by the provided probabilistic lower bound (see Fig.~\ref{fig:ctrl_ind}e), which empirically validates Proposition~\ref{prop:QP_MPC}. We provide demonstration videos for ACC experiments under both in-distribution (\videoInD) and OOD (\videoOOD) scenarios.
% ==================================================
\section{Conclusion and Future Work}\label{sec:conclusion}
This paper presents a method for state and uncertainty estimation from images using DNNs. In particular, we developed a Deep Ensemble for effective uncertainty quantification against adversarial attacks and OOD scenarios. Integrated with the Conformal Prediction, this approach provides formal coverage guarantees, which are demonstrated using a realistic traffic simulator. We also introduced Conformal Tube MPC to predict future state trajectories with statistical coverage, applying these constraints in ACC. Simulation results show effective car-following and safe distance maintenance. Future work will explore more complex autonomous driving applications.
% ==================================================

\clearpage
\bibliography{ref}
\end{document}